\documentclass[journal]{IEEEtran}
\usepackage[english]{babel}

\usepackage[letterpaper,top=2.01cm,bottom=1.52cm,left=1.69cm,right=1.69cm,marginparwidth=1.75cm]{geometry}

\usepackage{amsfonts}
\usepackage{amsmath, bm}
\usepackage{graphicx}
\usepackage{algorithm}
\usepackage[mathscr]{euscript}
\usepackage[noend]{algpseudocode}
\usepackage[colorlinks=true, allcolors=blue]{hyperref}
\usepackage{verbatim} 
\usepackage{multirow}
\usepackage{booktabs}
\usepackage{cite}
\usepackage{caption}
\usepackage{subcaption}

\usepackage{scrextend}

\title{4DenoiseNet: Adverse Weather Denoising from Adjacent Point Clouds}
\author{Alvari Seppänen, Risto Ojala, Kari Tammi}

\begin{document}
\maketitle

\begin{abstract}
Reliable point cloud data is essential for perception tasks \textit{e.g.} in robotics and autonomous driving applications. 
Adverse weather causes a specific type of noise to light detection and ranging (LiDAR) sensor data, which degrades the quality of the point clouds significantly.
To address this issue, this letter presents a novel point cloud adverse weather denoising deep learning algorithm (4DenoiseNet).
Our algorithm takes advantage of the time dimension unlike deep learning adverse weather denoising methods in the literature.
It performs about 10\% better in terms of intersection over union metric compared to the previous work and is more computationally efficient.
These results are achieved on our novel SnowyKITTI dataset, which has over 40000 adverse weather annotated point clouds.
Moreover, strong qualitative results on the Canadian Adverse Driving Conditions dataset indicate good generalizability to domain shifts and to different sensor intrinsics.
\end{abstract}


\section{Introduction}

Adverse weather conditions can have a huge impact on light detection and ranging (LiDAR) sensor data.
Airborne particles, such as rain droplets\cite{wallace2020full, goodin2019predicting, Bijelic_2020_STF}, fog \cite{bijelic2018benchmark, wallace2020full, heinzler2019weather, Bijelic_2020_STF}, or snowflakes \cite{jokela2019testing, kutila2020benchmarking, michaud2015towards, o1970visibility, Bijelic_2020_STF} cause undesired reflections, refractions, and absorptions of the laser, which results in missing and cluttered points.
This is a major problem since point clouds are often used for determining the open volume of the environment, \textit{e.g.} autonomous robots use point clouds for obstacle avoidance.
Moreover, the clutter also affects other downstream perception algorithms, such as object detection \cite{do2022lossdistillnet, do2022missvoxelnet, hahner2022lidar, hahner2021fog}, which are an essential component of autonomous road vehicles.
Thus, robust perception data in adverse weather is crucial as fatality rates for human drivers are notably higher in such conditions, as reported by the European Commission \cite{RoadsafetyintheEuropeanUnion} and the US Department of Transportation \cite{HowDoWeatherEventsImpactRoads}.

Our task is to remove points from LiDAR point clouds that are caused by airborne particles.
The motivation for this is to provide clean point cloud data for all downstream tasks \textit{e.g.} mapping, localization, object detection, and navigation.
Previous work uses either a classical approach \cite{charron2018noising, kurup2021dsor, park2020fast, wang2022scalable, li2022snowing} or a learned approach \cite{heinzler2020cnn}.
Unlike previous work, our work utilizes spatial-temporal data as an input to a neural network.
We do this by feeding adjacent point clouds $\mathbf{P}^{(t)} \in \mathbb{R}^{n\times 3}$ and $\mathbf{P}^{(t-1)} \in \mathbb{R}^{n\times 3}$ to a novel neural network that predicts the points that are caused by airborne particles.
Then the predictions are used for removing these points from the point cloud, yielding a clean point cloud $\mathbf{P}^{(t)'}$.
The neural network architecture can be optimized for multiple types of adverse weather \textit{e.g.} rain, fog, and snowfall.
Moreover, our algorithm is also tested with more challenging clutter caused by light, medium, and heavy snowfall, whereas previous work \cite{heinzler2020cnn} was tested only in fog and rain.
We obtained better performance than WeatherNet \cite{heinzler2020cnn} and general-purpose state-of-the-art semantic segmentation networks \cite{cortinhal2020salsanext, zhu2021cylindrical}.
Furthermore, the experiments show that our model generalizes better to other adverse weather conditions.

The problem of removing noise caused by adverse weather is not trivial because of the nature of the LiDAR sensor, sparsity of the point cloud, occlusions caused by the noise, and the varying density of the noise.
This leads to statistical or hard-coded filters \cite{charron2018noising, kurup2021dsor, park2020fast, wang2022scalable, li2022snowing} to remove valid points and preserve the clutter.
To address this problem, we use a deep learning architecture that learns an equivariance function.
A deep learning method is also able to utilize the complex patterns of adverse weather noise.
Due to the nature of the LiDAR sensor, the reflectance and position of solid structures affect the pattern of the noise.
Our method is trained with partly simulated semi-synthetic data.
Since labels are "free lunch" from the simulation, the training set can be built effortlessly, and our model can be trained in a supervised manner.
However, we want to emphasize that our model is tested quantitatively with real data captured in adverse weather.

We show that the key to robust adverse weather denoising is the efficient utilization of spatial-temporal data.
Spatial information, in metric space, is useful because clutter has relatively low density.
Temporal information is crucial because valid points follow predictable trajectories.
Contrarily, noise points caused by airborne particles have a chaotic nature.
Our architecture exploits these phenomena by a k-nearest neighbor convolution kernel that captures spatial-temporal information, and with a motion-guided attention mechanism.

Our contribution is two-fold.
\begin{itemize}
\item 
We present the first deep learning approach for LiDAR adverse weather denoising utilizing spatial and temporal information.
This is realized with a novel k-nearest neighbors search convolution on consecutive point clouds, which captures spatial and temporal information.
We surpass existing methods in performance by a large margin, with a lower computational cost.

\item
Since point-wise annotations are laborious, we train with semi-synthetic data generated by a highly realistic physics-based model \cite{hahner2022lidar}.
That is, synthetic effects of adverse weather are added to real point clouds captured in clear weather.
We are the first to use this data to train a model and test its performance on real data, captured in adverse weather.
The excellent performance indicates that our model is robust in this domain shift.
Given the excellent performance, our model will be an essential component in outdoor LiDAR sensor applications, enabling clean perception data for all downstream tasks.
Moreover, we present the first point-wise annotated adverse weather dataset, \textit{i.e.} SnowyKITTI, based on this simulation model which has approximately 40000 LiDAR scans.
\end{itemize}

\section{Related work}

Adverse weather denoising from sparse LiDAR point clouds is an emerging field, and only a handful of studies have been conducted.
Dense point cloud denoising is a more established field but the methods have not been tested for denoising adverse weather in sparse point clouds.
Therefore, we do not cover that area of research.

\subsection{Classical approaches}

Typically, the clutter caused by adverse weather has relatively low density.
Radius outlier removal (ROR) and statistical outlier removal (SOR), presented in \cite{rusu20113d}, remove outliers based on local density.
ROR removes points that do not have another point in distance $r$.
SOR computes the mean distance to $k$ nearest neighbors and decides if a point is an outlier based on the global mean distance between the points and the standard deviation.
These methods do not work well with point clouds that have inherently varying density, \textit{e.g.} LiDAR point clouds whose density is proportional to the measured range. 
This leads to the removal of distant points, and therefore these methods are not suited for adverse weather denoising.

Charron \textit{et al.} \cite{charron2018noising} proposed dynamic radius outlier removal (DROR) which adjusts the distance $r$ based on the range of the point from the sensor.
They achieved good results removing points caused by snowfall.
Dynamic statistical outlier removal (DSOR) \cite{kurup2021dsor} combines SOR and DROR.
It removes outliers based on a threshold that is defined by the global mean distance between the points and the standard deviation and the measured range of the point.
Low-intensity outlier removal (LIOR) \cite{park2020fast} is designed to remove points caused by snowfall.
The threshold is derived from LiDAR measured intensity, and it is a function of the measured range.
LIOR includes the ROR filter for preserving points that have low intensity but high density.
That is, LIOR takes advantage of the typical low intensity and density of airborne snowflakes.
However, valid points are removed from light-absorbent and semi-transparent surfaces.
Dynamic distance–intensity outlier removal (DDIOR) \cite{wang2022scalable} fuses DSOR and LIOR to achieve better performance.
Li \textit{et al.} utilized spatial-temporal features for removing points caused by snowfall \cite{li2022snowing}.
They showed that temporal information is valuable in this task by performing well compared to other methods.
Their method thresholds points to relative distances in W-T-space where W and T denote spatial dimensions and time, respectively.

\subsection{Learned approaches}

Since our method is based on segmentation with a neural network, we present related work from this area.
LiDAR point cloud semantic segmentation networks are focusing on general segmentation, and they can be trained to segment highly abstract shapes. 
Most successful approaches use voxelized \cite{tang2020searching, zhu2021cylindrical}, bird's eye view \cite{zhou2021panoptic}, or spherical projection input \cite{wu2018squeezeseg, wu2019squeezesegv2, xu2020squeezesegv3, razani2021lite}.
However, raw point input approaches exist as well \cite{qi2017pointnet, qi2017pointnet++, landrieu2018large, su2018splatnet, tatarchenko2018tangent, hu2020randla, rosu2022latticenet, yan2021sparse}
Cylinder3D \cite{zhu2021cylindrical} parts the point cloud into cylindrical voxels.
Then a 3D convolutional neural network extracts features and produces the predictions.
PolarNet \cite{zhou2021panoptic} uses bird's eye view pseudo image representation for feature extraction.
Spherical projection image input is beneficial for memory usage because of the dense representation \cite{cortinhal2020salsanext, milioto2019rangenet++}.
Another benefit of the projection image is that a 2D convolutional neural network can be used for feature extraction. 
However, 2D convolution kernels fail to capture local spatial information accurately due to the nature of the projection.
Work by Xu \textit{et al.} combines projection, voxel, and point inputs to produce more accurate segmentation results \cite{xu2021rpvnet}.

The semantic segmentation networks are well-proven and general-purpose.
Thus, they can be trained for segmenting the clutter caused by adverse weather. 
However, they require a lot of labeled training data, memory, and computational power due to the enormous amount of trainable parameters.
Recent work by Heinzler \textit{et al.} \cite{heinzler2020cnn} proposed the WeatherNet, an optimized semantic segmentation network for segmenting the clutter caused by fog and rain.
Their method has a significantly lower amount of trainable parameters while having equal or even better performance compared to the state-of-the-art general-purpose segmentation networks.
While their work is focused on segmenting the clutter caused by rain and fog, our method presents a novel spatial-temporal feature encoder and benchmarks the performance of light, medium, and heavy snowfall, which causes more severe clutter. 
Furthermore, our method has only 0.6M trainable parameters while theirs has 1.5M.
Therefore, our method generally requires less labeled data and generalizes better. 

\section{Methods}

\begin{figure*}[!t]
    \centering
    \includegraphics[width=0.96\textwidth]{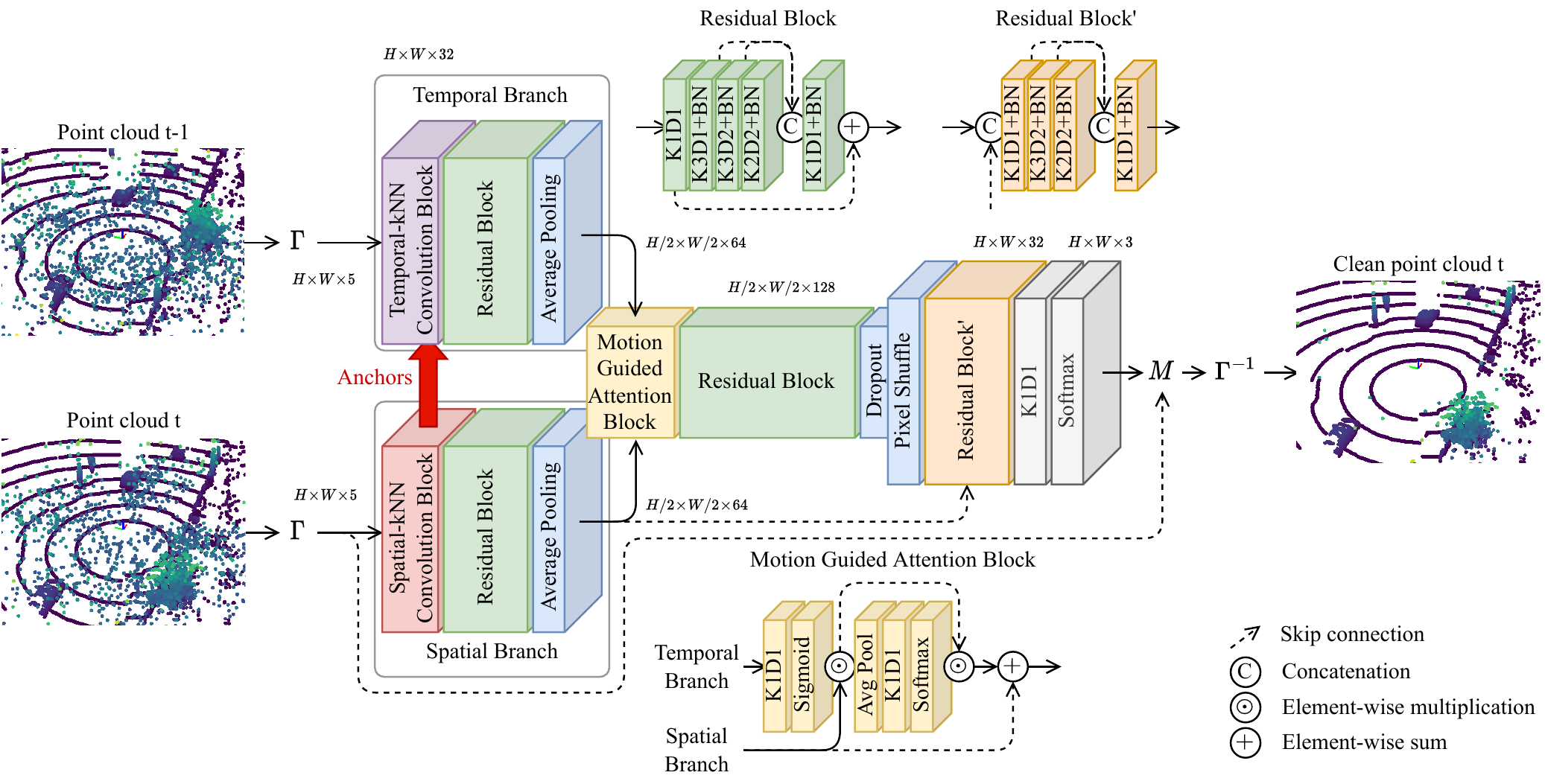}
    \caption{4DenoiseNet architecture. K, D, and BN indicate the kernel size, dilation, and batch normalization, respectively.}
    \label{fig:net}
\end{figure*}

\subsection{Ordered point cloud representation}

Point coordinates from a typical LiDAR sensor $\Vec{c} = (x, y, z)$ are mapped $\Gamma :\mathbb{R}^{n \times 3} \to \mathbb{R}^{s_h \times s_w \times 3}$ to spherical coordinates, and finally to image coordinates, as defined by

\begin{align}
    \begin{bmatrix}
        u \\
        v 
    \end{bmatrix}
    =
    \begin{bmatrix}
        1/2(1-\tan ^{-1}(yx^{-1})\pi ^{-1}) s_w \\
        (1- (\sin ^{-1}(z ||\Vec{c}||^{-1}) + f_{\mathit{vup}})f_v^{-1}) s_h
    \end{bmatrix}
\end{align}

\noindent where $(s_h, s_w)$ are the height and width of the desired projection image representation, $f_v$ is the total vertical field-of-view of the sensor, and $f_{vup}$ is the vertical field-of-view spanning upwards from the horizontal origin plane.
The resulting list of image coordinates is used to construct a $(x, y, z)$-channel image \textit{i.e.} ordered point cloud $\mathbf{P}_o \in \mathbb{R}^{s_h \times s_w \times (3+C_f)}$, where $C_f$ denotes the number of feature channels, in our case, $C_f=1$ for intensity.

\subsection{Utilization of spatial information}

\begin{figure}[!t]
    \centering
    \includegraphics[width=0.48\textwidth]{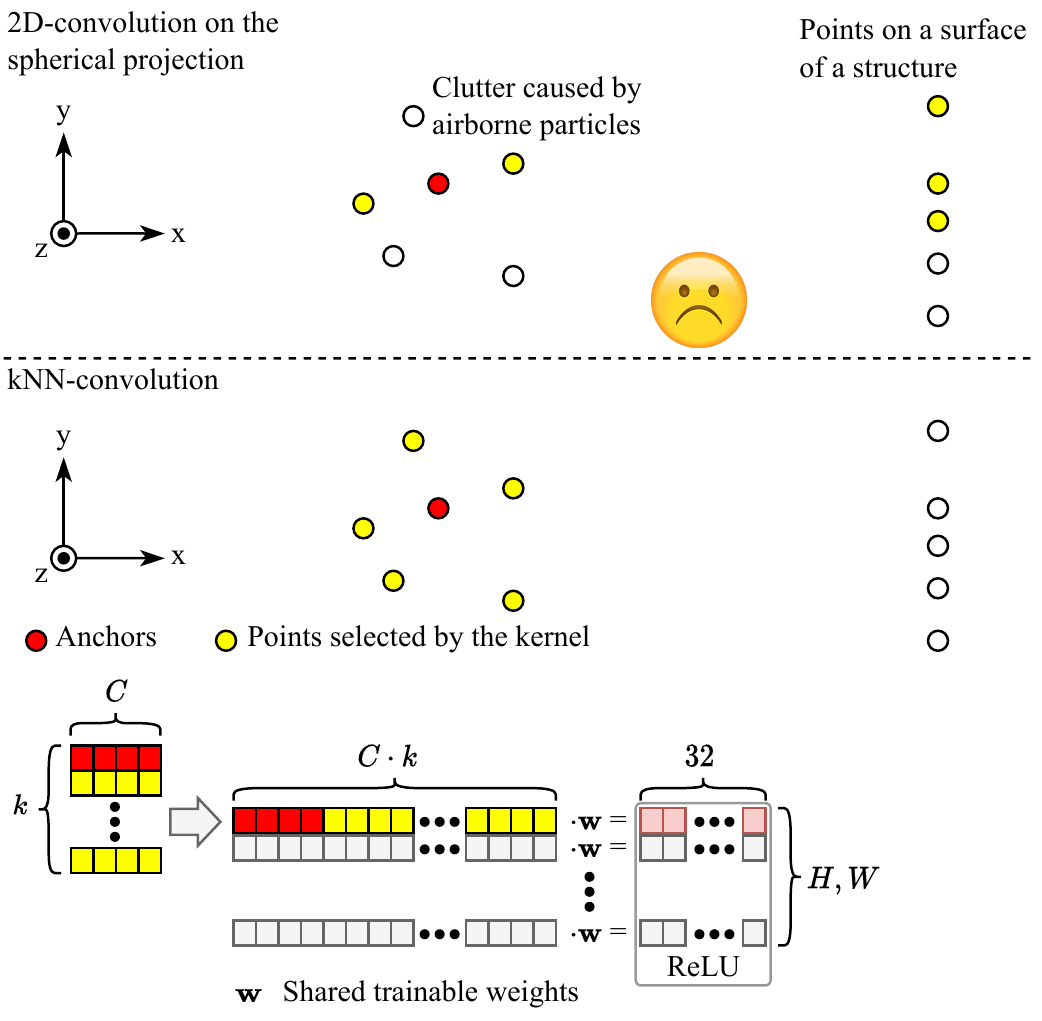}
    \caption{A highlight of an issue of a 2D-convolution on a spherical projection image. It fails to capture local points in metric space, whereas kNN-convolution captures local points, which is important in our task as the clutter is not continuous on the projection image.}
    \label{fig:knn_conv}
\end{figure}

Classical methods DROR \cite{charron2018noising}, DSOR \cite{kurup2021dsor}, and DDIOR \cite{wang2022scalable} show that local point density is a useful indicator for determining if a given point is caused by an airborne particle.
Therefore, enabling the neural network to capture this information is crucial.
Since our method is projection-based and a traditional convolution fails to capture local points \cite{sirohi2021efficientlps}, a measure has to be taken. 
We define the first convolution layer to capture k-nearest neighbors (kNN) in metric space via kNN-convolution.
An illustrative schematic is presented in Figure \ref{fig:knn_conv}. 
This convolution kernel considers the closest k points in the metric space instead of the neighboring points based on the pixel coordinates, enabling better spatial information for the network.
To mitigate the computational burden of the kNN-search, we search only in the neighboring area of the anchor point in the ordered point cloud.
We show an improvement compared to a traditional convolution in an ablation study (Section \ref{ablation_study}).
For simplicity, a pixel coordinate is denoted as

\begin{align}
    \Vec{p} = (u, v). 
\end{align}

\noindent The spatial-kNN-convolution becomes

\begin{align}
    \mathbf{\Theta}(\Vec{p}) = \mathbf{w} * \mathbf{P}_{o}^{(t)}  
    = \sum_{\partial \Vec{p}=0}^{k} \mathbf{w}(\partial \Vec{p}) \cdot \mathbf{P}_{o}^{(t)}(\psi(\Vec{p}))(\partial \Vec{p})
\end{align}

\noindent where $\mathbf{w}$ denotes trainable weights and $\psi(\Vec{p})$ is defined as 

\begin{align}
    \psi(\Vec{p}) 
    = \mathit{argmink}(|\mathbf{P}_{or}^{(t)}(\Vec{p} - \Vec{\xi}, ..., \Vec{p} + \Vec{\xi}) - \mathbf{P}_{or}^{(t)}(\Vec{p})|)
\end{align}

\noindent where $\mathit{argmink}(\cdot)$ returns indicies of minimum-k elements, and $\mathbf{P}_{or}^{(t)}$ denotes the range channel.
$\Vec{\xi}$ is a hyperparameter that defines the number of elements that are considered in the kNN-search.
Gathered kernel inputs are activated with a $\mathit{ReLU}(\mathbf{\Theta}(\Vec{p}))$ function.

\subsection{Utilization of temporal information}

The effects of adverse weather in LiDAR point clouds have a more chaotic nature than valid points.
This is caused by a reflection of the beam from an airborne particle.
The reflections caused by these particles are more unpredictable since they are small and are moved by turbulent airflow.
This chaotic behavior is on a much smaller scale in other points. 
Thus, temporal information can be utilized in our task. 
An empirical study shows that a reflection from an airborne particle, \textit{e.g.} snowflake, is extremely unlikely to be occurring twice in the same place.
That is, a single beam reflected from an airborne particle is highly unlikely to occur in the adjacent scan for a given beam, whereas reflections of other surfaces are more predictable.

\begin{figure}[!t]
    \centering
    \includegraphics[width=0.48\textwidth]{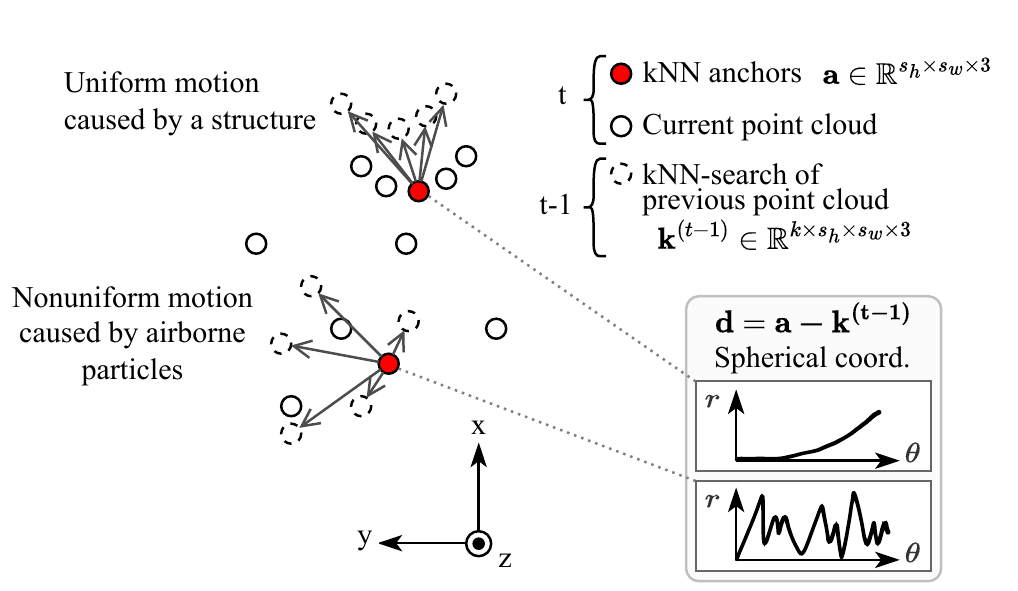}
    \caption{The Temporal-kNN-kernel capturing temporal information. Distributions given by $\mathbf{d} = \mathbf{a} - \mathbf{k}^{(t-1)}$, in the spherical coordinate system, are smoother for a set of points with a uniform motion. On the contrary, a set of points caused by a nonuniform motion are more random.}
    \label{fig:kNN-kernel}
\end{figure}

As illustrated in Figure \ref{fig:kNN-kernel}, we capture temporal information by searching for a kNN-set of points from a previous point cloud $\mathbf{P}^{(t-1)}$ using anchor points $\mathbf{a}\in \mathbb{R}^{s_h \times s_w \times 3}$ \textit{i.e.} the Cartesian channels of the current point cloud $\mathbf{P}^{(t)}$.
Similarly to the $\mathbf{P}^{(t)}$ kNN-search, the search considers only the neighboring area in the ordered point cloud.
The temporal-kNN-convolution is defined as

\begin{align}
    \mathbf{\Delta}(\Vec{p}) = \mathbf{w}_\Delta * \mathbf{d} \nonumber \\ 
    = \sum_{\partial \Vec{p}=0}^{k} \mathbf{w}_\Delta(\partial \Vec{p}) \cdot (\mathbf{a}(\Vec{p}) - \mathbf{P}_{o}^{(t-1)}(\psi_\Delta(\Vec{p})))(\partial \Vec{p})
\end{align}

\noindent where $\psi_{\Delta}(\Vec{p})$ is defined as 

\begin{align}
    \psi_{\Delta}(\Vec{p}) 
    = \mathit{argmink}(|\mathbf{P}_{or}^{(t-1)}(\Vec{p} - \Vec{\xi}, ..., \Vec{p} + \Vec{\xi}) - \mathbf{P}_{or}^{(t)}(\Vec{p})|).
\end{align}

\noindent $\mathbf{d} \in \mathbb{R}^{k\times s_h \times s_w \times 3}$ is an approximation of the motion of a local manifold.
$\mathbf{d}$ is converted from Cartesian into a spherical coordinate system to have a more relevant representation of the data, \textit{i.e.} $r$ and $(\theta, \phi)$ are the magnitude and the direction of the motion, respectively,

\begin{align}
    \Gamma_{\mathbf{d}} : 
    \begin{cases}
        \mathbb{R}^3 \to \mathbb{R}^3 \\
        (x,y,z) \mapsto (r,\theta,\phi),
    \end{cases}
\end{align}

\noindent this is done before convolving with $\mathbf{w}_\Delta$.
Similarly to the spatial-kNN-convolution, kernel output is activated with a $ReLU(\mathbf{\Delta}(\Vec{p}))$ function.

\subsection{Neural network design and training}

Based on the hypotheses, a function that captures both spatial and temporal features should have superior performance over its counterparts.
We approximate this function by a novel neural network, namely 4DenoiseNet  (Figure \ref{fig:net}), where 4D is a reference to time as the fourth dimension.
It has two branches: spatial and temporal.
The spatial branch processes the spatial features of the current point cloud $\mathbf{P}^{(t)}$, and the temporal branch processes the temporal features of the previous point cloud $\mathbf{P}^{(t-1)}$. 
The spatial branch has a kNN-convolution block that captures kNN for each point.
The kNN-search is also computed in the temporal branch.
However, anchors are provided by the spatial branch and subtracted from the kNN points of the previous point cloud $\mathbf{P}^{(t-1)}$.
This captures temporal information as illustrated in Figure \ref{fig:kNN-kernel}.
Both branches have a residual block for encoding, inspired by the work of Cortinhal \textit{et al.} \cite{cortinhal2020salsanext} and Aksoy \textit{et al.} \cite{aksoy2020salsanet}.
The branches are fused by the motion-guided attention (MGA) block, which fuses temporal features with spatial features using the motion-guided attention mechanism, described in \cite{li2019motion}. 
After another residual block, a pixel shuffle decreases channel dimension and increases spatial dimensions.
The residual block' connects the skip connection from the spatial branch with the main pipeline.
A standard convolution and Softmax normalization layers give us the desired logits.
Finally, a mask module $M$ removes points caused by airborne particles using the Softmax confidences, giving us a clean point cloud $\mathbf{P}^{(t)'}$.

Training objective is a standard cross-entropy and Lov\'asz-Softmax loss \cite{berman2018lovasz}, which optimizes for the Jaccard index \cite{jaccard1901etude} \textit{i.e.} intersection over union (IoU) metric

\begin{align}
    \mathscr{L}_{ls} = \frac{1}{|C|}\sum_{c \in C} \overline{\Delta_{J_c}} (m(c))
    && \text{and} \nonumber \\
    m_i(c) = 
    \begin{cases}
        1-x_i(c), & \text{if } c=y_i(c) \\
        x_i(c), & \text{otherwise}.
    \end{cases}
\end{align}

\noindent where the number of classes in denoted with $C$, $\overline{\Delta_{J_c}}$ defines the Lov\'asz extension of the Jaccard index, $y_i(c) \in \{ 0, 1\}$ and $x_i(c) \in [0, 1]$ hold the ground truth label and the prediction of pixel $i$ for class $c$, respectively.
The complete loss function is formulated as follows

\begin{align}
    \mathscr{L} = \mathscr{L}_{ls} + \mathscr{L}_{ce}
\end{align}

\noindent where $\mathscr{L}_{ce}$ denotes the standard cross-entropy loss.

\section{Experiments}

\subsection{Experimental setup} \label{experimental_setup}

\textbf{Datasets.} We conduct qualitative tests on the Canadian Adverse Driving Conditions dataset \cite{pitropov2021canadian}.
It is captured in real-world adverse weather conditions which include light, medium, heavy, and extreme snowfall.
Since this dataset does not have point-wise annotations, we train our model on our SnowyKITTI dataset.
SnowyKITTI is a modified KITTI odometry benchmark dataset \cite{geiger2012we} with synthetic snowfall created with a highly realistic physics-based simulation model presented in \cite{hahner2022lidar}. 
SnowyKITTI is divided into training, validation, and testing splits with a ratio of 40/10/50.
We also evaluate our model with the SnowyKITTI testing set.
Testing and validation sets are different sequences compared to the training set.
Furthermore, the training set is divided into subsets depending on the parameters of the snowfall simulation.
The subsets are described in Table \ref{datasets}.
This is done to investigate the generalizability of the models.
That is, a network is trained only with \textit{e.g.} light snowfall conditions and tested in all conditions. 
We have created a diverse selection of training subsets that validate robust performance across a wide range of adverse weather conditions.

\begin{table}[!t]
\centering
\caption{Definitions of different adverse weather conditions and training subsets.}
\label{datasets}
\begin{tabular}{ c | c c c }
\toprule
Classification    & Light        & Medium       & Heavy        \\
\midrule
\midrule
Snowfall rate     & $[0.5, 1.5[$ & $[1.5, 2.5[$ & $[2.5, 3.0]$ \\
Terminal velocity & $[1.0, 2.0]$ & $[1.0, 2.0]$ & $[1.0, 2.0]$ \\
\midrule
All               & \checkmark   & \checkmark   & \checkmark   \\
Subset 1          & \checkmark   & \checkmark   & -            \\
Subset 2          & \checkmark   & -            & \checkmark   \\
Subset 3          & -            & \checkmark   & \checkmark   \\
Subset 4          & -            & -            & \checkmark   \\
Subset 5          & -            & \checkmark   & -            \\
Subset 6          & \checkmark   & -            & -            \\
\bottomrule
\end{tabular}
\end{table}

\textbf{Evaluation metrics.} We use the Jaccard index \cite{jaccard1901etude} \textit{i.e.} IoU, which is formulated as follows

\begin{align}
    J_c(\mathbf{y}, \mathbf{x}) = \frac{|\{ \mathbf{y}=c\} \cap \{ \mathbf{x}=c\}|}{|\{ \mathbf{y}=c\} \cup \{ \mathbf{x}=c\} |}.
\end{align}

\noindent where $\mathbf{y}$ and $\mathbf{x}$ denote ground truth and predicted labels, respectively.
We are only interested in the IoU value of the noise class.
Thus, from here forward IoU is for noise class.

\textbf{Training and inference details.} The parameters of the network are optimized with the Adam optimizer \cite{kingma2014adam}.
We fix the spatial dropout probability to 0.2.
The initial learning rate is set to 0.01 which is decayed by 0.01 after each epoch.
An L2 penalty is applied with $\lambda = 1 \cdot 10^4$ and a momentum of 0.9.
Data is augmented by randomly dropping points before creating the projection, applying a random translation and rotation, and flipping randomly relative to the y-axis.
These augmentation methods are applied independently of each other with a probability of 0.5.
Inference runtime is benchmarked with an Nvidia GTX 1060 GPU.

The code and the data are made publically available here \footnote{\label{github} https://github.com/alvariseppanen/4DenoiseNet}.

\subsection{Quantitative results}

The main results are presented in Table \ref{performance}.
Our 4DenoiseNet is compared with the other adverse weather filtering model WeatherNet \cite{heinzler2020cnn}, and with general-purpose state-of-the-art semantic segmentation networks \cite{cortinhal2020salsanext, zhu2021cylindrical}.
The models are evaluated in terms of Jaccard index, runtime, and parameter count in Light, Medium, and Heavy snowfall.
Furthermore, the models are trained with different subsets (Table \ref{datasets}) to investigate generalizability to conditions differing from training conditions.
All learned models in Table \ref{performance} are trained and tested with all conditions, albeit training and testing do not share any sequences.
The training set is 80\% of the size of the testing set as described in Section \ref{experimental_setup}.
Overall our model performs the best in both accuracy and runtime.
Moreover, our model has fewer trainable parameters compared to the other learned models.
There is a large gap in accuracy between classical DROR and LIOR compared to learned approaches, highlighting the difficulty of this task.

\begin{table}[!t]
\centering
\caption{Trained with all conditions. GPU: Nvidia GTX 1060, CPU: 4.0 GHz Intel i5-7600K 5th generation. * - no training required. Bolded font indicates the best values.}
\label{performance}
\begin{tabular}{ c | c c c | c c }
 \toprule
 \multirow{2}{3em}{Method} & \multicolumn{1}{c}{Light} & \multicolumn{1}{c}{Medium} & \multicolumn{1}{c}{Heavy} & \multicolumn{1}{|c}{Runtime} & \multicolumn{1}{c}{Param.} \\
 & IoU & IoU & IoU & ms & $\cdot 10^6$ \\ 
 \midrule
 \midrule
 DROR* \cite{charron2018noising}
 & 0.442 & 0.445 & 0.437 & 120  & $10^{-5}$\\
 LIOR* \cite{park2020fast}
 & 0.445 & 0.442 & 0.430 & 120  & $10^{-5}$\\
 \midrule
 SalsaNext \cite{cortinhal2020salsanext}
 & 0.947 & 0.947 & 0.951 & 26   & 8.8  \\
 Cylinder3D \cite{zhu2021cylindrical}
 & 0.949 & 0.943 & 0.942 & 65  & 41.7  \\
 WeatherNet \cite{heinzler2020cnn}
 & 0.884 & 0.889 & 0.865 & 44   & 1.5  \\
 \midrule
 4DenoiseNet
 & \textbf{0.975} & \textbf{0.976} & \textbf{0.977} & \textbf{19}    & \textbf{0.6}  \\
\bottomrule
\end{tabular}
\end{table}

\begin{figure*}
     \centering
     \begin{subfigure}[b]{0.32\textwidth}
         \centering
         \includegraphics[width=1\textwidth]{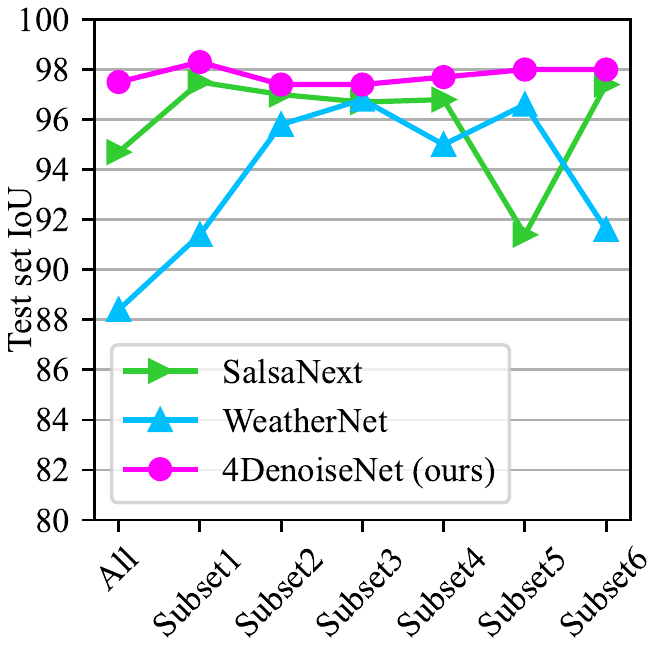}
         \caption{Light}
         \label{fig:light}
     \end{subfigure}
     \hfill
     \begin{subfigure}[b]{0.32\textwidth}
         \centering
         \includegraphics[width=1\textwidth]{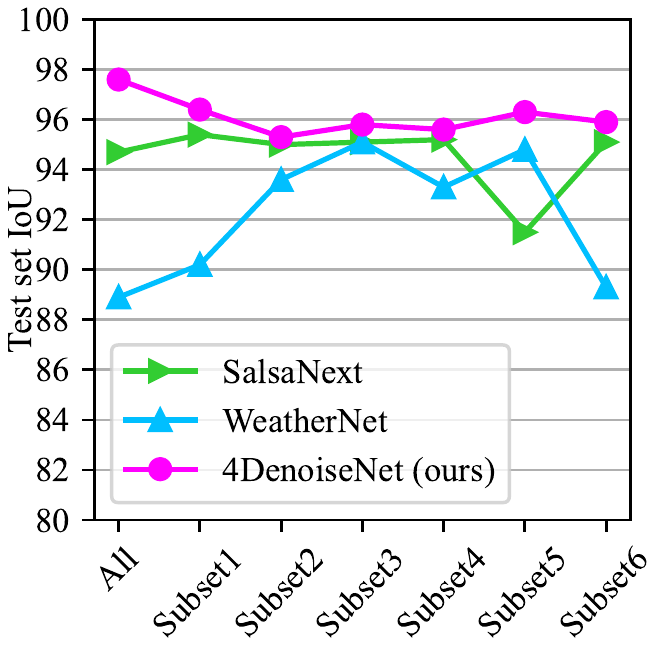}
         \caption{Medium}
         \label{fig:medium}
     \end{subfigure}
     \hfill
     \begin{subfigure}[b]{0.32\textwidth}
         \centering
         \includegraphics[width=1\textwidth]{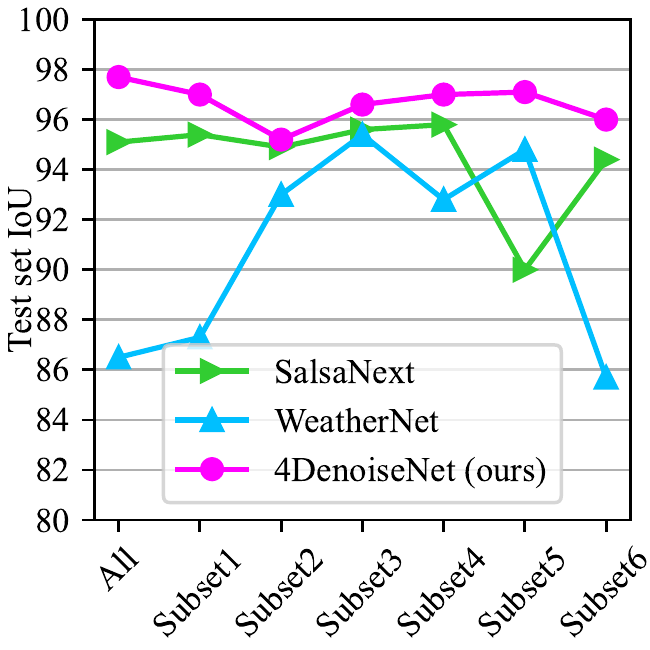}
         \caption{Heavy}
         \label{fig:heavy}
     \end{subfigure}
        \caption{Quantitative results on the Light, Medium, and Heavy test sets with different training sets (Table \ref{datasets}). The performance of 4DenoiseNet remains high regardless of the training dataset.}
        \label{fig:performance_degradation}
\end{figure*}

Figure \ref{fig:performance_degradation} presents the performance when models are trained with different training subsets.
It should be noted, that the testing sets are separate from the training sets. 
In the figure, the x-axis presents which training set was used.
Most notably, our model is the most robust as there is a smaller variance in IoU compared to other models. 
When trained with Subset2, SalsaNext is on par with our model.
We hypothesize that it is more unlikely to overfit with Subset2 because it has more variance in conditions.
This hypothesis is supported by the performance with Subset5 where the performance of SalsaNext is lower.
When the other models are trained with all conditions, their performance is lower, this might be caused by conditions imbalance of the dataset, which causes the models to overfit to the most frequent condition.
However, our model performs the best when trained with all conditions in Medium and Heavy snowfall.

\subsection{Ablation study} \label{ablation_study}

To study the importance of spatial-temporal kNN-convolution, we conducted an ablation study where four 4DenoiseNet variants were compared.
Table \ref{ablation_table} describes the variants and their performance, runtime, and the number of trainable parameters. 
The kNN-convolution module was replaced with a traditional 2D convolution layer.
In the variants without the temporal branch, MGA is not used, and the spatial branch connects straight to the second residual block.
Based on the ablation study, the spatial kNN-convolution increases the performance slightly more than the temporal kNN-convolution.
Overall, both spatial and temporal kNN-convolution modules improved the performance significantly.

\begin{table}[!t]
\centering
\caption{Ablations of different modules and their contribution to the IoU and runtime. Our kNN-convolution module is replaced with a traditional 2-dimensional convolution (First conv. 2D) on the ordered point cloud. \checkmark indicates that previous point cloud $\mathbf{P}_{t-1}$ and the temporal branch are used.}
\label{ablation_table}
\begin{tabular}{ c c | c c c | c c }
 \toprule
 \multirow{2}{2em}{\centering First\\conv.} & \multirow{2}{2em}{\centering $\mathbf{P}_{t-1}$} & \multicolumn{1}{c}{Light} & \multicolumn{1}{c}{Medium} & \multicolumn{1}{c}{Heavy} & \multicolumn{1}{|c}{Runtime} & \multicolumn{1}{c}{Param.} \\
 & & IoU & IoU & IoU & ms & $\cdot 10^3$ \\ 
 \midrule
 \midrule
 2D    & -          & 0.599 & 0.566 & 0.562 & 8    & 477.84 \\
 2D    & \checkmark & 0.676 & 0.654 & 0.666 & 13   & 568.92 \\
 kNN   & -          & 0.882 & 0.864 & 0.866 & 16   & 480.53 \\
 \midrule
 kNN   & \checkmark & 0.975 & 0.976 & 0.977 & 19   & 571.61 \\
\bottomrule
\end{tabular}
\end{table}

\subsection{Qualitative results and discussion}\label{Q_R_and_D}

\begin{figure*}
     \centering
     \begin{subfigure}[b]{0.32\textwidth}
         \centering
         \frame{\includegraphics[width=1\textwidth]{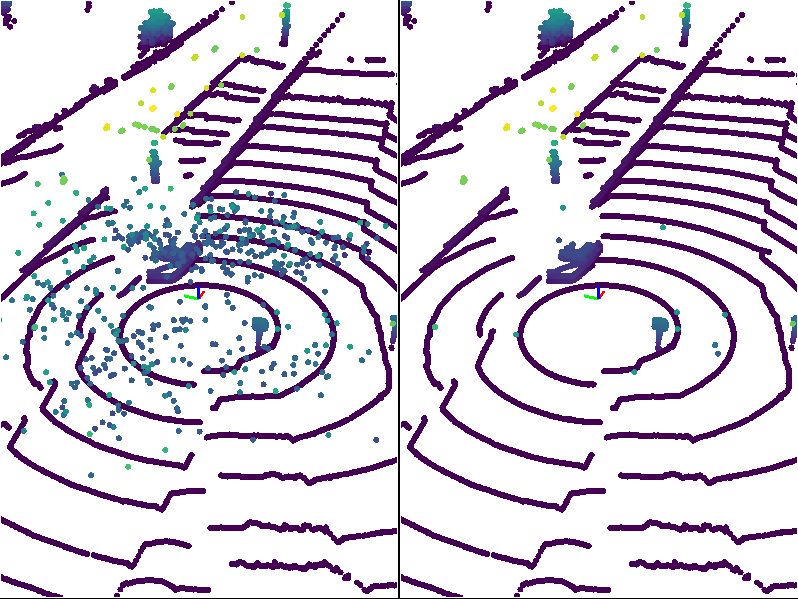}}
         
         \vspace{0.2cm}
         
         \frame{\includegraphics[width=1\textwidth]{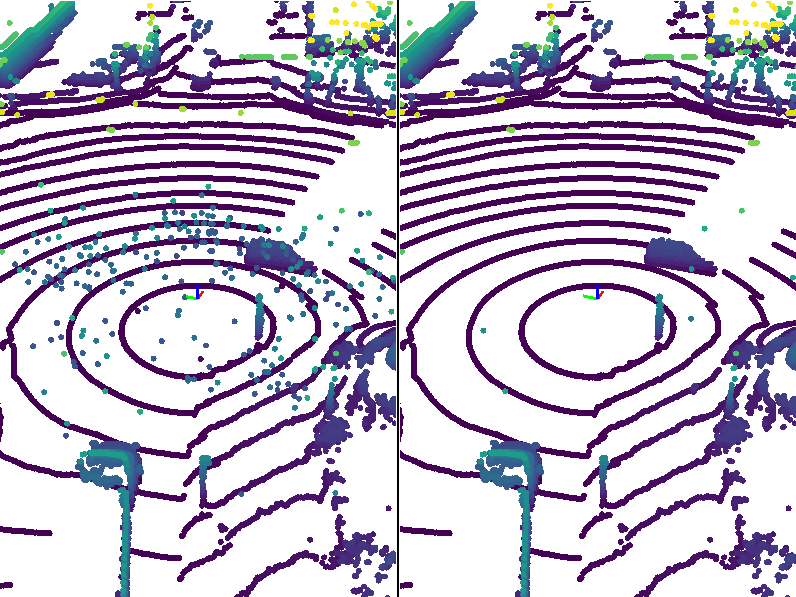}}
         \caption{Light}
         \label{fig:light_real}
     \end{subfigure}
     \hfill
     \begin{subfigure}[b]{0.32\textwidth}
         \centering
         \frame{\includegraphics[width=1\textwidth]{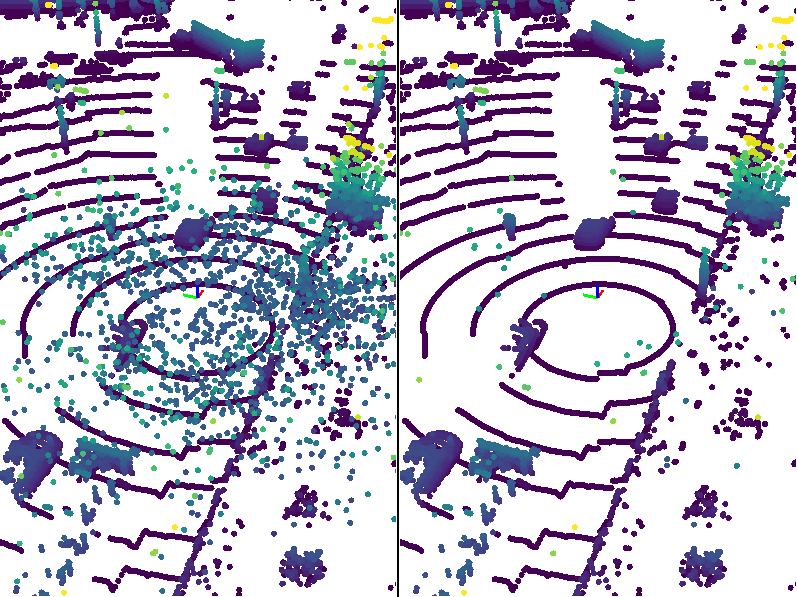}}
         
         \vspace{0.2cm}
         
         \frame{\includegraphics[width=1\textwidth]{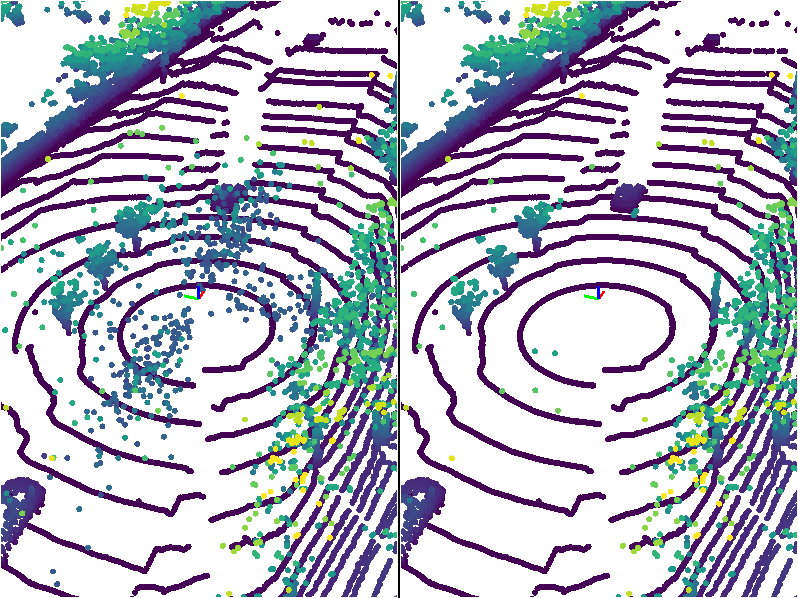}}
         \caption{Medium}
         \label{fig:medium_real}
     \end{subfigure}
     \hfill
     \begin{subfigure}[b]{0.32\textwidth}
         \centering
         \frame{\includegraphics[width=1\textwidth]{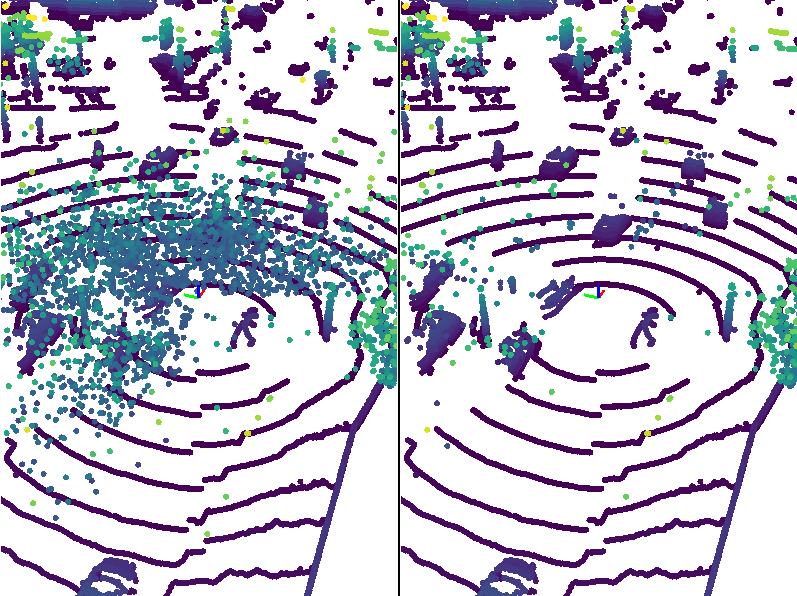}}
         
         \vspace{0.2cm}
         
         \frame{\includegraphics[width=1\textwidth]{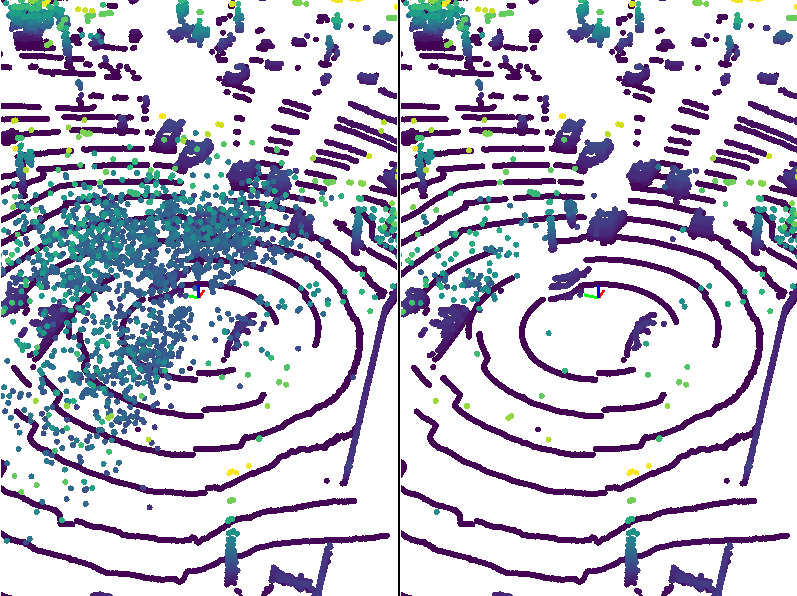}}
         \caption{Heavy}
         \label{fig:heavy_real}
     \end{subfigure}
        \caption{Denoising performance on 6 individual sample point clouds from real snowfall. Raw input and denoised output are on the left and right sides of each image pair, respectively. We suggest zooming in for better detail.}
        \label{fig:realdata}
\end{figure*}

We conducted qualitative tests on real LiDAR data captured in adverse weather.
The data is from the Canadian Adverse Driving Conditions dataset \cite{pitropov2021canadian}.
The performance was analyzed in snowfall as it causes more noise to the point cloud compared to other conditions.
Figure \ref{fig:realdata} illustrates the denoising performance.
The reader should focus on the framed image pairs, where the image on the left side is the raw input, and the image on the right side is the denoised output.
In medium and heavy snowfall, some objects are not visible in the input point cloud, but after applying our algorithm they are clearly visible.
Based on visual analysis, our model does not seem to remove valid points albeit being sparse (see the first Medium pair).

To further highlight the challenge of this task, the LiDAR used from this dataset has different intrinsic parameters compared to the LiDAR used for training.
The training LiDAR has a vertical resolution of 0.44$^\circ$. 
The testing LiDAR has a vertical resolution of 1.25$^\circ$. 
This means that the training point clouds are denser compared to the testing point clouds.
Furthermore, the performance on the domain shift improves when the intensity channel is omitted.
The trained models that we tested (Table \ref{performance}) get biased on the intensity.
Therefore, we omitted the intensity channel and trained our model with $(r, x, y, z)$-input channels.
After this, the model generalizes well to the real data.

\section{Conclusions}

We presented 4DenoiseNet, the first deep learning algorithm for adverse weather denoising on adjacent LiDAR point clouds.
Quantitive results on our annotated SnowyKITTI dataset, qualitative results on the Canadian Adverse Driving Conditions dataset, and runtime indicate that our model is the new state-of-the-art on adverse weather denoising.
4DenoiseNet is based on the spatial-temporal kNN-convolution module that is presented in this work. 
We show via ablation study the importance of spatial and temporal information in metric space in adverse weather denoising tasks.
Given the light computational demand and the performance, our algorithm will be an essential component in outdoor LiDAR applications, enabling clean point cloud data for all downstream tasks.
For more qualitative results we refer to our repository page \footref{github}.


\bibliographystyle{unsrt}
\bibliography{bibliography}

\end{document}